\documentclass[]{spie}  

\usepackage{amsmath,amsfonts,amssymb}
\usepackage{graphicx}
\usepackage[colorlinks=true, allcolors=blue]{hyperref}

\title{SPIE Proceedings: Style template and guidelines for authors}

\author{Zhengyi Luo, Austin Small, Liam Dugan, Stephen Lane \\
Department of Computer and Information Science, University of Pennsylvania
}

\usepackage{cite}
\usepackage{amsmath,amssymb,amsfonts}
\usepackage{graphicx}
\usepackage{multirow}
\usepackage{textcomp}
\usepackage{xcolor}
\usepackage{array}
\usepackage{tabu}
\usepackage{amsmath}
\usepackage{algorithm}
\usepackage[noend]{algpseudocode}
\graphicspath{ {images/} }
 
\begin{document} 

\title{Cloud Chaser: Real Time Deep Learning Computer Vision on Low Computing Power Devices
}

\onecolumn 
\maketitle
\begin{abstract}
Internet of Things(IoT) devices, mobile phones, and robotic systems are often denied the power of deep learning algorithms due to their limited computing power. However, to provide time critical services such as emergency response, home assistance, surveillance, etc, these devices often need real time analysis of their camera data. This paper strives to offer a viable approach to integrate high-performance deep learning based computer vision algorithms with low-resource and low-power devices by leveraging the computing power of the cloud. By offloading the computation work to the cloud, no dedicated hardware is needed to enable deep neural networks on existing low computing power devices. A Raspberry Pi based robot, Cloud Chaser, is built to demonstrate the power of using cloud computing to perform real time vision tasks. Furthermore, to reduce latency and improve real time performance, compression algorithms are proposed and evaluated for streaming real-time video frames to the cloud.\\


\end{abstract}

\section{Introduction}
With the growing success of deep learning in the field of computer vision, it is natural for developers and researchers alike to be interested in deploying these new vision algorithms in devices such as smart home cameras, robots, drones, etc. However, popular computer vision algorithms that leverage deep neural nets often require high end GPUs (Graphics Processing Units) to achieve desired performance. This constraint heavily limits the number of algorithms researchers can experiment with. To overcome this problem, researchers have tried different approaches, such as designing more efficient deep learning networks like MobileNet \cite{HowardZCKWWAA17}, making computing add-on modules like Intel Movidius \cite{Intel}, or creating dedicated chips and processors such as NeuPro by CEVA \cite{NeuPro} 

The idea of using Cloud infrastructure to aid low resource devices, is first introduced by James Kuffner \cite{Kuffner2012}, has inspired a number of research projects \cite{Hu2012} \cite{Goldberg2013}. As more and more Cloud Computing services provide deep learning frameworks, deep learning algorithms are becoming more accessible, especially when GPU enabled cloud machines provided by Amazon AWS, Google Cloud, Paperspace etc. are offering substantial computing power at affordable prices. Moreover, these cloud machines tend to have relatively high-bandwidth networks, making them suitable for real-time applications. For instance, the Nvidia Gaming as a Service (GaaS)\cite{cloudgaming} offers players who do not have access to high-end GPUs the opportunity to run their games in the cloud and render them on their personal devices. In this paper, we are interested in using these resources to enable low resource devices to run computationally intensive real-time computer vision algorithms. Specifically, object detection is chosen as the task for investigation. 

As more and more internet of things, robotics, and mobile devices become equipped with cameras, object detection will be increasingly important as it is the foundation on which higher level tasks such as navigation, planning, and surveillance can be built. State of the art object detection algorithms all heavily rely on powerful GPUs to achieve a desirable accuracy and frame-rate. In order to achieve similar performance in low computing power devices, we offload the computing tasks to a GPU enabled cloud instance. 

This paper makes the following contributions: 

\begin{itemize}
    \item Proposing a software architecture to allow resource constrained devices to run state of the art deep learning object detection algorithms that require a GPU to achieve real time performance.
    \item Utilizing multi-thread and asynchronous programming to coordinate real-time video streaming and object detection between cloud and local devices.
    \item Designing and evaluating compression algorithms to reduce latency induced by offloading computing work to the cloud. 
\end{itemize}

The paper is organized as follows: section II reviews the current research in the field of deep learning for object detection and other efforts in running deep neural nets on low resource devices. Section III describes the research objective and problem formulations. Section IV gives the system architecture, and Section V will describe our technical approach and methodology. System experiments and performance analysis will be in section VI, while section VII concludes the paper.

\section{Related Works}

\subsection{Deep Learning Object Detection}
Object Detection, different from image classification, is the task of detecting possible objects in the current image frame, as well as producing a bounding box that indicates the location of the object. State of the art object detection algorithms consists of two main approaches: region Based, such Regional-based convolutional network (R-CNN)\cite{GirshickDDM13} and Fast-RCNNs \cite{Girshick15},  and single-shot based You only look once (YOLO) \cite{Redmon2015YouDetection} and Single shot multibox Detector (SSD) \cite{Liu2016SSD:Detector}. For regional proposal based approaches, they often consist of two steps: regional proposal and object classification. At first, the potential objects' locations (bounding box) in the current scene are formulated, then a deep neural network such as a Convolutional Neural Network (CNN) are used to predict the objects' classes. R-CNN, \cite{GirshickDDM13} was the first model to adopt this procedure for object detection. Fast R-CNN \cite{Girshick15} speeds up the R-CNN model by using a single network for the whole image rather than dedicating a separate one to each region. Faster R-CNN \cite{Ren2017} brings in near real-time performance at 5 FPS frame rate on a GPU. However, the separation of the object detection problem into both a regional proposal stage and classification stage results in complicated pipelines that significantly slow down the algorithm. 

YOLO, a state of the art object detection algorithm that was introduced in 2015, pioneered the approach to unify the steps of regional proposal and object classification \cite{RedmonDGF15}. Due to its simple and effective architecture, YOLO can achieve comparable accuracy at more than 30 FPS. Since YOLO was first released, several improvements in speed and accuracy have since been realized in YoloV3 \cite{abs-1804-02767}.

The advent of these real time object detection algorithms has encouraged numerous researchers to attempt to integrate them into real-time systems. However, all of these algorithms rely on powerful GPUs to achieve real-time performance. For instance, to achieve real time performance, the YOLO algorithm requires 4 GB of GPU Random-access memory (RAM), which is only available on higher gaming PCs and Laptops. On typical resource constraint devices such as mobile phones and IoT devices, most of the time discrete GPU is unavailable.

\subsection{Real Time Computer Vision On Resource Constraint Devices}
Several Neural Network based approaches have been devised to cater towards real time object detection on resource constrained platforms.  Many of these approaches either compress a pre-trained network like \cite{Wang2017FactorizedCN} or directly train a small network such as \cite{IandolaMAHDK16}. For instance, MobileNet proposes the use of depth-wise separable convolutions in order to reduce computation and model size by reducing the complexity of the convolution operation \cite{HowardZCKWWAA17}.  While these works do reduce space and computational complexity, a significant trade off in accuracy is present. 

Though there are a few online machine learning service provider like \cite{google} \cite{amazon} \cite{Clarifai}, existing services focus on image analysis and object detection for business purposes. These services mainly depends on deriving business insight derived from image analysis, so they do not provide object detection in a desired frame rate that is suitable for real time tasks.

With state of the art object detection unable to run in real time on low resource platforms without compromising on accuracy, an opportunity exists to offload the computing workload to the cloud. Existing work attempts to measure and contrast the performance of state of the art object detection algorithms as it pertains to real time object tracking with aerial drones \cite{Lee2017RealTimeCO}.  While this work can act as a proof of concept, practical concerns such as selection of communication protocols, image compression, and threading of tasks on the mobile platform to minimize communication latency all require significant investigation before cloud computing becomes a viable mainstream solution for real time object detection.

\section{Research Objective and Problem Formulation}
To enable deep learning based computer vision on resource constrained devices, we leverage the computing power of the cloud. Using the cloud as a secondary computing unit has many advantages. First of all, cloud computing allows any device with basic WiFi capabilities to run computationally intensive deep learning neural networks. This is especially useful as popular high performance vision libraries often need a significant amount of GPU power to produce satisfying results. Another advantage is the fact that incorporating a cloud computing component requires minimal additional setup but can drastically improve the device's capabilities. The only hardware requirement for using cloud computing in an existing device will be a reliable WiFi module. 

Thus, our goal is to provide a way to integrate deep learning computer vision algorithms into low resource devices with minimal effort using cloud technologies. We are assuming the constraint of a computing unit that has the capability of communicating through WiFi but has limited bandwidth and little to no on-board computing power with which to run highly parallel deep learning algorithms. To offload real time computer vision tasks to the cloud, a low-latency communication architecture is designed, implemented, and tested. In order to test the performance of the system, we have selected object detection tasks as the main objective for this work.

\section{System Architecture}
 
 \begin{figure*}[h!]
 \centering
  \includegraphics[scale=0.5]{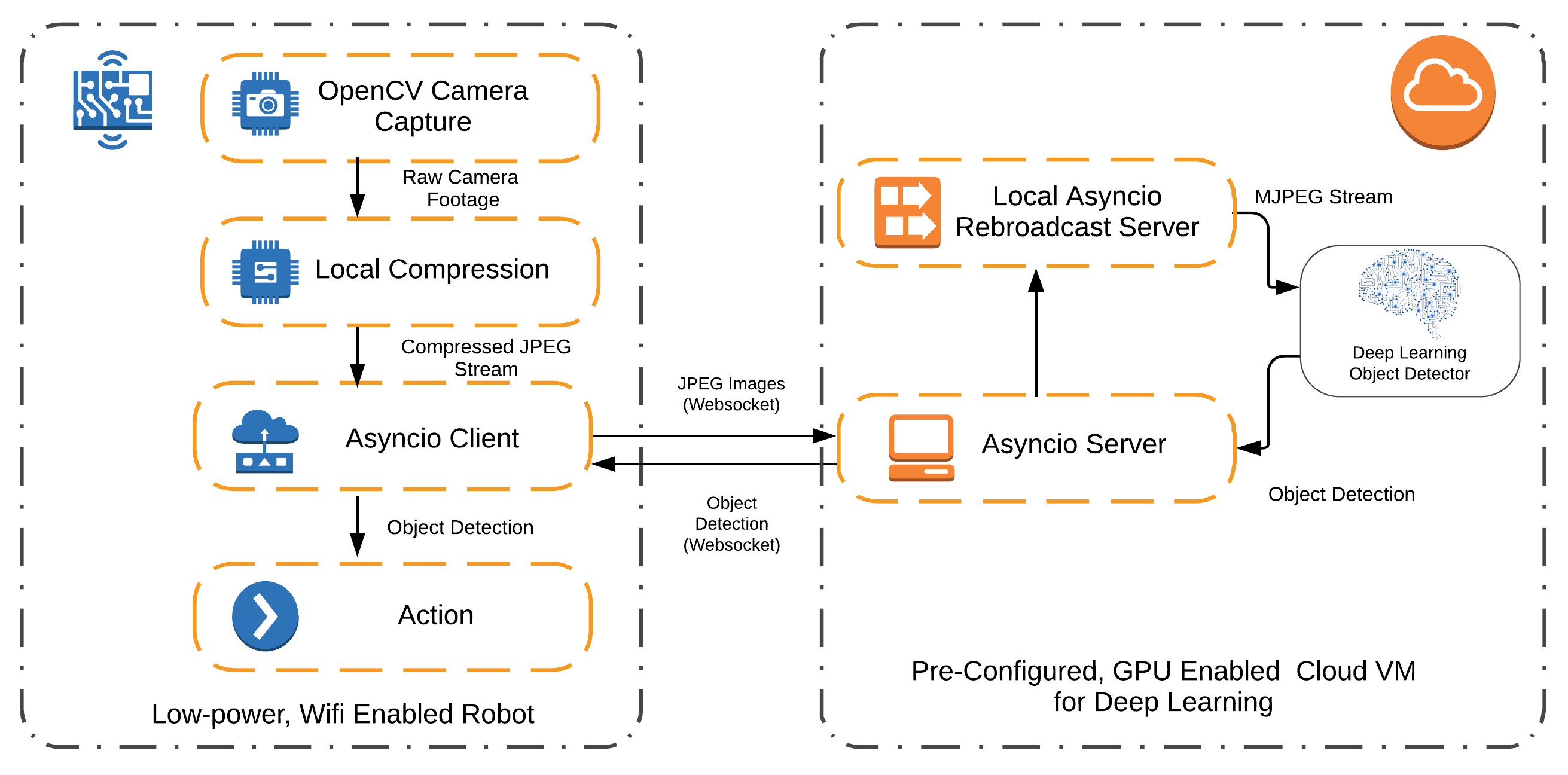}
  \caption{System Architecture}
  \label{fig:arch}
\end{figure*}

To enable offloading the heavy computation tasks to the cloud, a novel architecture is proposed, as shown in Fig. \ref{fig:arch}. The architecture consists of two main parts: the client side, which resides on the mobile device and handles the image data transmission, and the server side, which runs on the cloud instance and executes the object detection algorithms.

\subsubsection{Server Side }
The server side is made of three main components, the Asyncio Server, the Local Asyncio Rebroadcast Server, and the Deep Learning Object Detector. 

\begin{itemize}
\item Asyncio Server: runs two separate websocket servers for communicating with the client. The first server receives the incoming JPEG stream from the device, while the second server reads the results from the object detector and sends data to the client device. 
\item Local Asyncio Rebroadcast Server: receives the JPEG stream from the Asyncio Server and rebroadcasts the images using MJPEG protocol. 
\item Deep Learning Object Detector: analyzes the rebroadcast MJPEG frames, and uses YOLO-based deep learning object detection algorithms to recognize objects in the current frame.
\end{itemize}

\subsubsection{Client Side}
The client side is made of four main components, the OpenCV Camera Capture component, the on device image compression component (Local Compression), the Asyncio Client component, and the device Action component. 

\begin{itemize}
    \item OpenCV Camera Capture: uses OpenCV library to capture image output from the device camera.
    \item Local Compression: processes raw image data from camera and compresses it to reduce file size, then encodes the raw image data into JPEG images. 
    \item Asyncio Client: runs two separate websocket clients and communicates with the server. The first client sends the image data to the server continuously, and the second receives object detection data.
    \item Action: reacts to the detected object in the scene. 
\end{itemize}
\subsubsection{Pipeline Workflow Overview}
Our proposed pipeline contains the above components and uses multiple threads to ensure parallel execution. Each image frame captured by OpenCV Camera Capture from the device camera is compressed by Local Compression. Then the compressed image is encoded in JPEG format and sent to the server by the websocket through the Asyncio Client. After the cloud Asyncio Server receives the image frame, it shares the image with Local Asyncio Rebroadcast Server for rebroadcasting the frame as a MJPEG stream. The Deep Learning Object Detector accepts the image frames and detect the objects in it. After detection is finished, the cloud Asyncio Server reads it from the object detection algorithm, and sends the detection as a string through websocket to the client. The device Asyncio Client receives the object information and sends it to the Action component for reaction.

\section{Technical Approach}

\subsection{Real Time Video Streaming and Compression}
To stream real time video frames to the cloud from a local device to the cloud for analysis, we propose the following approaches to the corresponding challenges:

\subsubsection{Communication Protocol}
To enable real-time performance, the communication protocol needs to be low level enough that it introduces minimal latency. Specifically, we chose websocket as our main communication protocol. Another option to ensure the lowest possible network latency is raw TCP, but the direct use of TCP has limited speed advantage. Websocket allows multiple connections to a single server, and is easy to use in an existing web framework like the python aiohttp library. Since websocket ensures easy interfacing with the rest of the application, while also providing a more secure interface for the streamed video frames (WSS protocol encrypts streams using HTTPS) it was the best choice for our use case. 
    
\subsubsection{Network Address Translation}
In order to stream a MJPEG video stream through the network, existing programs usually set up a HTTP web server on local devices and access the video stream through the device's assigned IP address. However, for security reasons, Internet of Things devices and mobile devices are often assigned private IP addresses that are unreachable from outside of the private network. Naturally, a cloud machine is outside of the local network, thus making traditional streaming schemes unusable for our purposes.

Some users with administrative privileges on their routers can circumvent this problem through enabling port forwarding, which is a router setting that allows a user to open up specific ports on their routers to allow public access. However, this approach is not possible on public networks where common users do not have administrative privileges. Additionally, some routers do not support port forwarding. 

To make the device's video stream accessible from the cloud machine, as well as to avoid running a separate MJPEG server on our device, our system proposes the following solutions: instead of setting up a MJPEG server directly on the device, camera frames captured from the device are sent to the server via an established websocket connection between server and device. Local devices, though not accessible from public networks, can establish bi-directional websocket connections with the server. Thus, the client can send video frames through the websocket directly to the server, and the server can then rebroadcast the frames using the MJPEG protocol, which is then consumed by object detection algorithms. 
    	
\subsubsection{Network Latency and Compression}
Due to the limited bandwidth of WiFi networks, we propose several compression approaches in order to reduce the size of the data transmitted and reduce latency.
\begin{itemize}
    \item Camera resolution: by reducing the camera capturing resolution from 960 * 540 to 480 * 270, we cut down the number of pixels transmitted by a factor of 4.
    \item Blurring the image: by applying a blur effect on the image, we improve the JPEG compression algorithm's efficiency, and effectively cut down the data size transmitted. 
    \item Blacking out: by using the temporal data from the previously processed frame, we are able to identify known objects and black out all regions of the image which do not contain objects. This approach assumes that objects do not change their relative position drastically between frames, and by leaving a buffer region on the object bounding box (15px margin), the detection algorithm can still identify the existing objects in the new frame. To detect new objects that may appear in the current scene, a complete, not blacked out frame is periodically sent for detecting new objects. File size reduction and latency reduction from the above approaches are outlined in the experiment section. 
\end{itemize}

\subsection{Asynchronous and Multi-threaded Computing}
In our architecture, the local device needs to send a video stream continuously while receiving detection results from previous frames. The cloud service also receives and rebroadcasts each frame to localhost for deep learning object detection algorithms. On top of that, the cloud needs to start the object detection process on demand and read results from the process. To achieve these goals, asynchronous programming and multi-threading are heavily utilized in our system.

On the cloud/server side, three different threads run in parallel. One thread runs a websocket server for receiving the video stream.  Within this thread, the rebroadcast server asynchronously broadcasts the video stream, exposing it as a MJPEG stream on localhost. These two processes communicate through shared variables. A second thread runs a websocket server which starts the deep learning object detector as the third thread. This websocket server then reads the output from the deep learning object detector, and sends it back to the client. 

On the client/device side, three different threads are also initiated to ensure parallel computing. The first thread runs a websocket client dedicated to capturing, compressing, and sending a continuous real-time video stream to the server. The second thread also hosts a websocket client that receives the object detection result from the server in a non-ending loop. Finally, a third thread reads the information from the second websocket thread, and reacts to the results from the cloud object detector.

\subsection{Deep Learning Object Detection}
Deep learning based object detection algorithms have gained significant traction due to their rapidly improving performance on some of the most well-known image classification datasets such as ImageNet and COCO. For this work, we mainly focus on adapting the YOLO\cite{RedmonDGF15} object detection algorithm to a cloud computing friendly configuration, which is shown as the Deep Learning Object Detector in Fig. \ref{fig:arch}. 

Taking an input frame from the MJPEG stream provided by the Local Asyncio Rebroadcast Server as shown in Fig. \ref{fig:arch}, the image frame will first be resized to fit the network parameters, and then go through the following calculations to predict the current objects and their location in the scene. The problem formulation is as follows. 
\begin{itemize}
    \item [1)] An image is broken up into an SxS grid of cells (S = 7 in our case), where each cell is responsible for producing B (B = 5) bounding boxes. A bounding box represents a potential object in the scene. If the center of an object lies within a cell, that cell is responsible for generating the object's bounding box.
    
    \item[2)] Each cell predicts the conditional probability $Pr(Class_i | Object)$ which represents the probability that the object in a grid cell is of $Class_i$, given that an object is  present in the cell.
    
    \item [3)] For each bounding box, five predictions, (x, y, w, h, c) are produced, where (x, y) represents the center of the predicted bounding box of an object, (w, h) represents the width and height of the bounding box, and c represents the confidence. The confidence score c is calculated as formula (\ref{eq:1}), where c represents object class specific confidence in a bounding box, meaning how likely a bounding box is to contain class i as well as how well the bounding box fits the object.
    
\begin{equation} \label{eq:1}
\begin{split}
c &= Pr(Class_i | Object) * Pr(Object) * IOU_{pred}^{truth}  \\
&= Pr(Class_i) * IOU_{pred}^{truth}
\end{split}
\end{equation}
 
 where IOU stands for Intersection over Union, which is the area of overlap divided by the area of union between the detected bounding box and the ground truth bounding box.

\end{itemize}

\begin{figure}[t]
  \centering
  \includegraphics[scale=0.5]{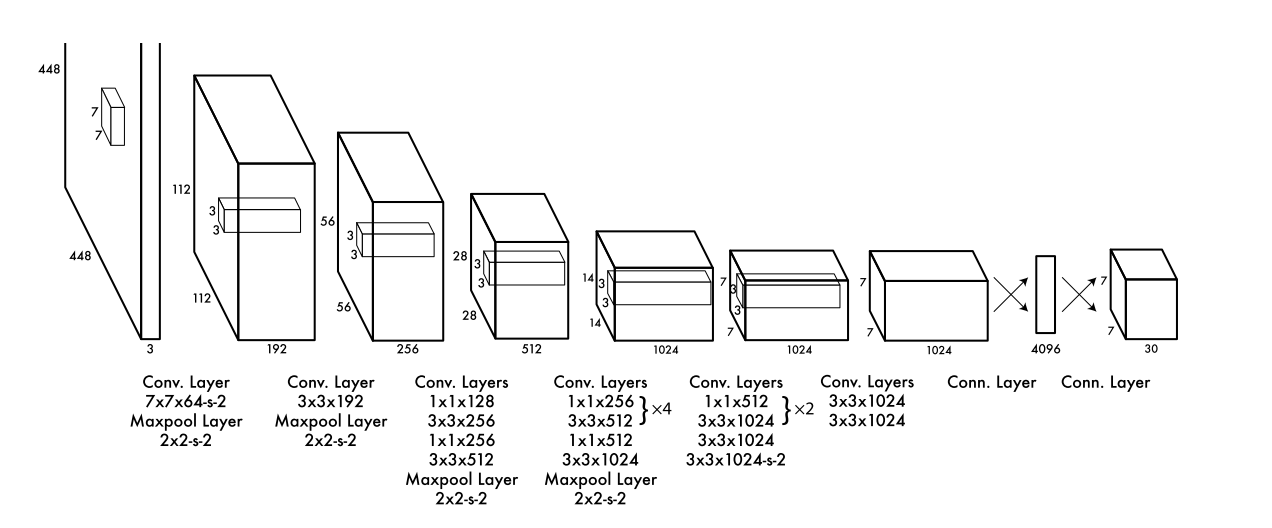}
  \caption{Neural Network Architecture for Deep Learning Object Detector}
  \label{fig:yolo_arch}
\end{figure}

An image frame goes through a Convolutional Neural Network (CNN) shown in Fig. \ref{fig:yolo_arch} to predict the (x, y, w, h, c) as formulated above. The neural network contains 24 convolutional layers and then 2 fully connected layers. This network was inspired by GoogleLeNet model for image classification \cite{Szegedy2015}. Unlike GoogLeNet, $1 \times 1$ reduction layers followed by $3 \times 3$ convolutional layers are used. The output of the CNN is then fed into to the Asyncio Server, where it will be transferred to the client.

For our application, the Deep Learning Object Detector is trained on the Microsoft COCO dataset, which contains pictures of 80 objects. The object detector is run as a subprocess of the server side program, as shown in Fig. \ref{fig:yolo_arch}, and the output of the detector is framed as (x, y, c, w, h). The confidence threshold is set at 50$\%$. Essentially, for each detected object in each frame that has a confidence greater than 50$\%$, the subprocess will output a text line giving its prediction. 

The advantage of using this construction lies in its speed and accuracy. By unifying the step of regional proposal and object classification, the final classification can be computed by one pass of the network, making the algorithm applicable to real-time applications with state of the art accuracy. Combining the high performance of YOLO with our cloud computing system, resource constrained devices can make use of this object detection system with minimal additional setup or hardware.

\section{Experiments}

Cloud Chaser \cite{cloudchaser}, shown in Fig. \ref{fig:chase}. was built to demonstrate the viability of our approach. It is our custom Raspberry Pi based robot capable of following voice commands, recognizing 80 different kinds of common objects in real time, and tracking these objects in real-time (thus the name "Cloud Chaser").

Using Cloud Chaser, three sets of experiment are conducted to demonstrate the viability of our approach. First, the communication latency introduced by offloading critical computation to the cloud is timed and calculated. Further, we timed the compression algorithms that are intended to reduce communication latency. Three compression algorithms are considered: averaging the captured image without reducing resolution, reducing captured image resolution, and blacking out parts of the image that are considered "uninteresting". Last but not least, to demonstrate the adaptability of our platform, an iOS app is developed to showcase our architecture running on different devices. 

\subsection{Experiment Settings}
\begin{figure}[t]
  \centering
  \includegraphics[scale=0.3]{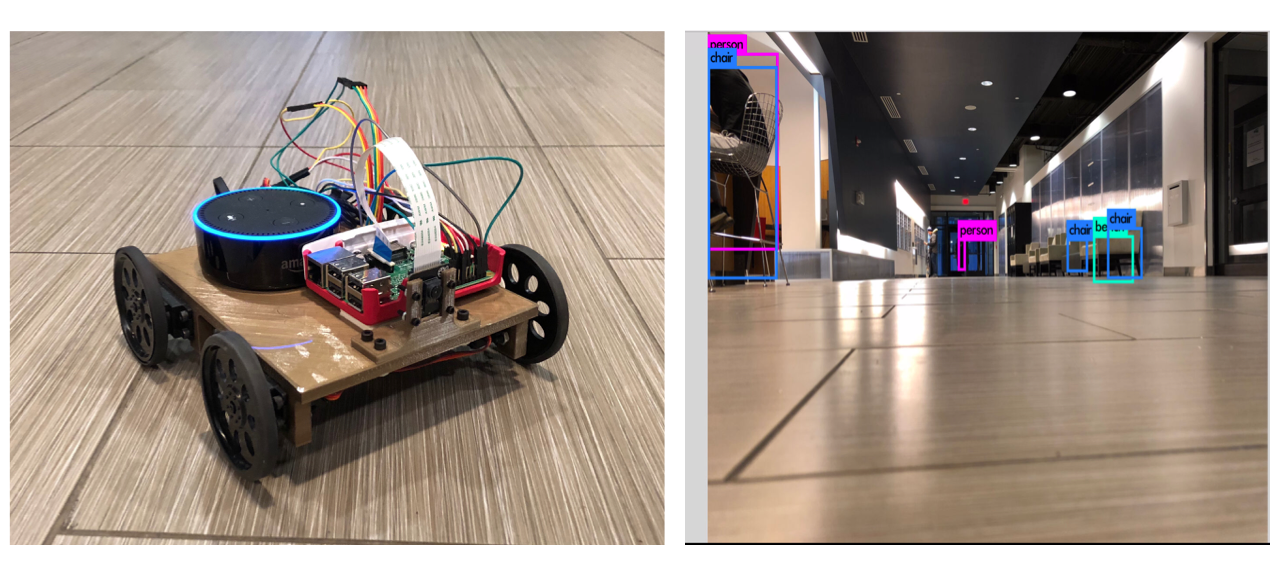}
  \caption{Cloud Chaser and cloud based real time object detection. The right image showcases a screenshot of real time objects being recognized from the Paperspace cloud instance.}
  \label{fig:chase}
\end{figure}

Our cloud instance is a Paperspace GPU+ instance, equipped with Nvidia Quadro P4000. Cloud Chaser, our custom built robot shown in Fig. \ref{fig:chase}, is built from a Raspberry Pi 3 B.

\subsection{Real-Time Communication Latency}
The latency induced by enabling our program can be broken into three parts: latency introduced by traveling through physical space when sending packets, latency introduced by limited bandwidth when sending data to cloud instance and sending back prediction, and the time it took for YOLO to process the image and produce prediction. We tested these latencies separately and in combination.

\begin{table}[t]
\centering
\caption{Time for round trip websocket communication, data size 26 bytes, taking the average of 2000 round trips}
\renewcommand{\arraystretch}{2}
\begin{tabu} to 0.8\textwidth { | X[c] | X[c] | X[c] | }
 \hline
  Timing Difference & Sunnyvale to San Francisco (41.5 mi) & Sunnyvale to New York (2,937.8 mi) \\
 \hline
 TCP & 0:00:00.023089  & 0:00:00.092086  \\
\hline
Websocket  & 0:00:00.024619  & 0:00:00.094308  \\
\hline
\end{tabu}
\label{table:1}
\end{table}

\begin{table}[b]
\centering
\caption{Time for round trip websocket communication, data size 13500 bytes, taking the average of 2000 round trips}
\renewcommand{\arraystretch}{2}
\begin{tabu} to 0.8\textwidth { | X[c] | X[c] | X[c] | }
\hline
  Timing Difference & Sunnyvale to San Francisco (41.5 mi) & Sunnyvale to New York (2,937.8 mi) \\
 \hline
Websocket  & 0:00:00.076612  & 0:00:00.210735  \\
\hline
\end{tabu}
\label{table:2}
\end{table}

\begin{table}[t]
\centering
\caption{Time between frame of person showing up and detection}
\renewcommand{\arraystretch}{2}
\begin{tabu} to 0.8\textwidth { | X[c] | X[c] |  X[c] | X[c] | }
 \hline
 Time  & Total Delay & Time  & Total Delay \\
 \hline
8:00 AM &  0:00:00.562203  & 6:00 PM &   0:00:00.651300\\
\hline
11:00 AM &  0:00:00.687919 & 9:00 PM &  0:00:00.756316 \\
\hline
3:00 PM &   0:00:00.585435 & 12:00 AM &  0:00:00.494050 \\
\hline
\end{tabu}
\label{table:3}
\end{table}

\begin{figure*}[t]
    \centering
  \includegraphics[scale=0.5]{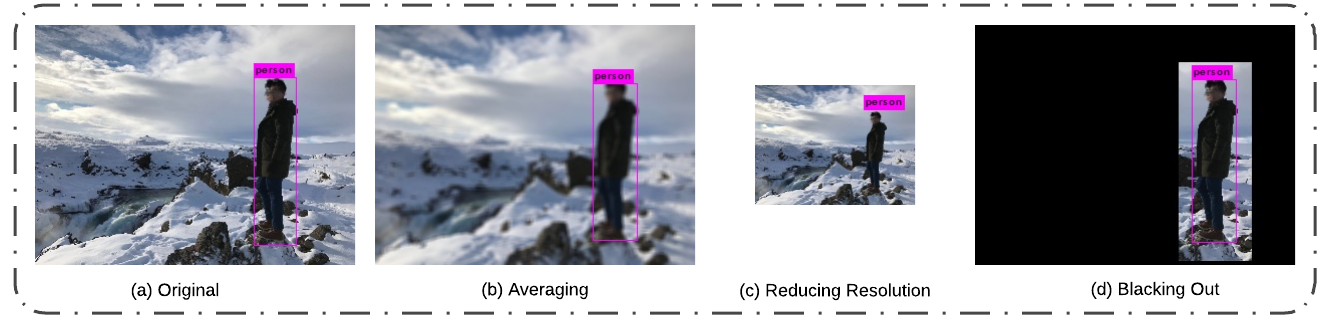}
  \caption{Prediction By Different Compression Approaches}
  \label{fig:compression}
\end{figure*}

\begin{table}[b]
\centering
\caption{Image resolution 480 * 270, file size in bytes}
\renewcommand{\arraystretch}{2}
\begin{tabu} to 0.8\textwidth { | X[c] | X[c] | X[c] | }
\hline
    & Compressed (averaging) & Original \\
 \hline
Latency  & 0:00:00.522536  & 0:00:00.651300  \\
\hline
Average File Size  & 14989 & 17740  \\
\hline
\end{tabu}
\label{table:4}
\end{table}

Table \ref{table:1} makes clear the time for a small packet to be sent from the local device (Raspberry Pi) to the server by calculating the round trip delay. From the data we conclude that using websocket against TCP introduces negligible additional delay(1ms) compared to the delay introduced by YOLO running at 15 FPS (60ms). 

To time the latency introduced by limited bandwidth, we encoded a string of length 13500 bytes, which is the average size of a 480 * 270 byte JPEG image. The delay introduced by sending this data packet through the network is shown in Table \ref{table:2}.

To time the combined latency of object detection and network communication, a picture of a person shown  in Fig. \ref{fig:compression}(a). is used as a token. First the camera is facing a black plane where no object is detected in the scene. After a few seconds the program swaps out the camera feed for the JPEG picture which is read from memory (counting the time to recompress the picture into JPEG format) and sent for detection. This effectively simulates the procedure of capturing, sending, and detecting a person in the current scene. The time when YOLO first detects the presence of a person is recorded. The results from five different times of day are shown in Table \ref{table:3}. While YOLO can run at an average frame rate of 15 FPS on our machine, the additional delay time between the picture event occurring on the local machine and cloud detection is caused by the following factors: 1) Network Congestion, where the cloud instance can not process the image frames sent from local devices as quickly as the local devices can send them, and 2) The video compression and decompression time taken by OpenCV on both the cloud instance and local device. Overall, by offloading the object detection from the local device, less than half of a second delay is introduced on the original performance of the object detection algorithm. Equipped with better GPU enabled cloud instance, the latency can be reduced even further.

\subsection{Compression Algorithms Evaluation}

In order to reduce the bottle-neck of transmission, we tried a few different compression schemes to reduce the amount of data that needed to be transmitted.

\subsubsection{Averaging}
Applying an averaging filter to the picture has the effect of reducing the sharpness of the picture, as shown in Fig.\ref{fig:compression}(b). The effect on file size and latency reduction of applying an averaging filter with each pixel weighted by 0.04 and averaging every 5 by 5 grids is shown in Table \ref{table:4}. Even though the image remains the same resolution after averaging, a blurred image will benefit more from JPEG compression and will thus take less data to transmit.

As a side effect of blurring this image we observe a reduction in accuracy on the our object detector. Our experiment showed that blurring using this kernel will reduce the mAP (Mean Average Precision) for YOLOv3 from 0.7465 to 0.6426 on the 2007 Pascal VOC dataset.

\begin{table}[t]
\centering
\caption{Image resolution 960 * 540 vs 480 * 270, file size in bytes}
\renewcommand{\arraystretch}{2}
\begin{tabu} to 0.8\textwidth { | X[c] | X[c] | X[c] | }
\hline
    & 960 * 540 & 480 * 270 \\
 \hline
Latency  & 0:00:00.522536  & 0:00:00.651300  \\
\hline
Average File Size  & 73198 & 17740  \\
\hline
\end{tabu}
\label{table:5}
\end{table}

\begin{table}[t]
\centering
\caption{Image size original vs blacked out 480 * 270, file size in bytes}
\renewcommand{\arraystretch}{2}
\begin{tabu} to 0.8\textwidth { | X[c] | X[c] | X[c] | }
\hline
    & Original & Blacked Out \\
 \hline
Latency  &  0:00:00.651300  & 0:00:00.651300  \\
\hline
Average File Size  & 17740 & 14418  \\
\hline
\end{tabu}
\label{table:6}
\end{table}

\subsubsection{Reduction on Resolution}
Reducing the resolution from 960 * 540 to 480 * 270 results in both a file size reduction and latency reduction as shown in Table \ref{table:5}. The affect of this reduction is shown in Fig.\ref{fig:compression}(c).

\subsubsection{Blacking out}

The size reduction effect of blacking out an image is shown in Table \ref{table:6}. As shown in Fig. \ref{fig:compression}(d), blacking out the uninteresting part of the image still results in the correct detection of the object in the current frame, assuming the location of objects do not change drastically between frames. 

\subsection{Platform Adaption}

To test the viability of this approach on other platforms, an iOS app is developed. Combined with the spacial mapping capability of the ARkit, the app demonstrates the ability to identify and label objects in a scene in real time. To demonstrate the real time object localization capability of the app, please refer to our short video \cite{chaseios}. 

Through these experiments, we have demonstrated that using the GPU-enabled cloud machines to carry out the deep learning object detection tasks is a viable, easy to setup, and easy to adapt approach. A variety of devices and platforms will be able to implement this architecture and make use of state of the art deep leaning object detection algorithms in real time.

\section{Conclusion and Future Work}

In this work, we present a novel architecture to integrate existing deep learning computer vision algorithm to resource constraint devices. As opposed to designing a new neural network that can be run on resource constraint devices, this approach runs the existing state of the art neural networks in the cloud and utilizes the wireless networking capabilities of the device to ensure real time performance. Since the full neural network is run in the cloud, there is little compromise on accuracy. Also, by utilizing multi-threading and asynchronous computing, coupled with compression algorithms for image size reduction, the latency introduced by streaming data to the cloud is reduced to minimum. Finally, as demonstrated in our experiment, devices with camera and WiFi capability can be easily adapted to use our approach for deep learning computer vision, allowing rapid prototyping for researchers and developers alike. 

In the future, we hope to develop a more robust communication scheme for streaming camera data from the device to the cloud. Also, we look forward to adapting more deep learning algorithms to this system, allowing resource constrained devices to utilize real time deep learning for a greater range of tasks. 

\bibliographystyle{unsrt}
\bibliography{mendeley}

\end{document}